\documentclass[sigconf, nonacm=true]{acmart}

\usepackage{epsfig}
\usepackage{graphicx}
\usepackage[american]{babel}
\usepackage{microtype}
 
\usepackage{xcolor}

\usepackage{color, colortbl}
\usepackage{multirow}
\usepackage{subfigure}
\usepackage{booktabs}
\usepackage{comment}

\newcommand{\etc}{\textit{etc.}}
\newcommand{\eg}{\textit{e.g.}}
\newcommand{\Eg}{\textit{E.g.}}
\newcommand{\etal}{\textit{et al.}}
\newcommand{\ie}{\textit{i.e.}}

\begin{document}

\title{\textit{Face.evoLVe}: A High-Performance Face Recognition Library}


\author{Qingzhong Wang}
\authornotemark[1]
\email{wangqingzhong@baidu.com}
\affiliation{%
  \institution{Baidu Research}
  \city{Beijing}
  \country{China}
}

\author{Pengfei Zhang}
\authornote{Equally contribute to this work.}
\email{pengfeizhang0520@gmail.com}
\affiliation{%
  \institution{National University of Defense Technology}
  \city{Changsha}
  \country{China}
}

\author{Haoyi Xiong}
\authornotemark[2]
\email{xionghaoyi@baidu.com}
\affiliation{%
  \institution{Baidu Research}
  \city{Beijing}
  \country{China}
}

\author{Jian Zhao}
\authornote{Corresponding authors: Jian Zhao (Homepage: \url{https://zhaoj9014.github.io}) and Haoyi Xiong.}
\email{zhaojian90@u.nus.edu}
\affiliation{%
  \institution{Institute of North Electronic Equipment}
  \city{Beijing}
  \country{China}
}

\begin{abstract}
   While face recognition has drawn much attention, a large number of algorithms and models have been proposed with applications to daily life, such as authentication for mobile payments, etc. Recently, deep learning methods have dominated in the field of face recognition with advantages in comparisons to conventional approaches and even the human perception. Despite the popular adoption of deep learning-based methods to the field, researchers and engineers frequently need to reproduce existing algorithms with unified implementations (\emph{\ie, the identical deep learning framework with standard implementations of operators and trainers}) and compare the performance of face recognition methods under fair settings (\emph{\ie, the same set of evaluation metrics and preparation of datasets with tricks on/off}), so as to ensure the reproducibility of experiments.

   %
   %
   
   To the end, we develop \underline{face.evoLVe} --- a comprehensive library that collects and implements a wide range of popular deep learning-based methods for face recognition. First of all, face.evoLVe is composed of key components that cover the full process of face analytics, including face alignment, data processing, various backbones, losses, and alternatives with bags of tricks for improving performance. Later, face.evoLVe supports multi-GPU training on top of different deep learning platforms, such as PyTorch and PaddlePaddle, which facilitates researchers to work on both large-scale datasets with millions of images and and low-shot counterparts with limited well-annotated data. More importantly, along with face.evoLVe, images before \& after alignment in the common benchmark datasets are released with source codes and trained models provided. All these efforts lower the technical burdens in reproducing the existing methods for comparison, while users of our library could focus on developing advanced approaches more efficiently.
   %
   %
   Last but not least, face.evoLVe is well designed and vibrantly evolving, so that new face recognition approaches can be easily plugged into our framework. Note that we have used face.evoLVe to participate in a number of face recognition competitions and secured the first place. 
   The version that supports PyTorch \cite{paszke2019pytorch} is publicly available at \url{https://github.com/ZhaoJ9014/face.evoLVe.PyTorch} and the PaddlePaddle \cite{ma2019paddlepaddle} version is available at \url{https://github.com/ZhaoJ9014/face.evoLVe.PyTorch/tree/master/paddle}.  Face.evoLVe has been widely used for face analytics, receiving 2.4K stars and 622 forks.
\end{abstract}



\maketitle

\begin{figure}[t]
    \centering
    \includegraphics[width=\linewidth]{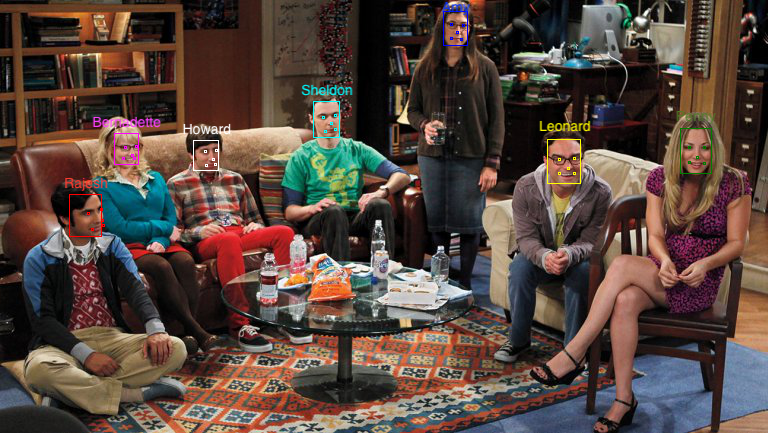}\\
    \includegraphics[width=\linewidth]{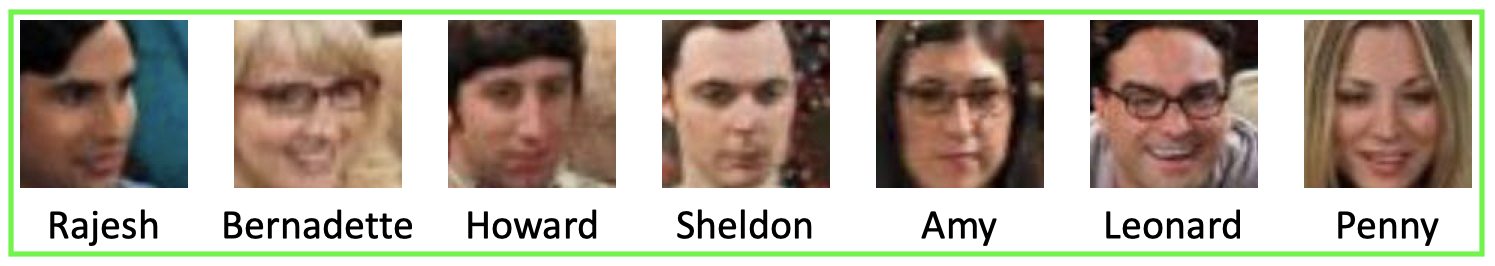}
    \caption{Face analytics from a scene of ``The Big Bang Theory'' TV series.}
    \label{fr-example}
\end{figure}

\section{Introduction}

As a critical research problem in multimedia computing~\cite{zhu2011multimedia}, face recognition has drawn much attention from both academia and industry with intensive applications to daily life, 
%
\eg, authentications for mobile payment \cite{du2018mobile} and the friends tagging from photos/videos (\eg, Fig. \ref{fr-example}). A face recognition system normally takes an image or a video as input and identifies faces in the image or video as outputs. Recently, deep learning-based approaches have dominated in the field of face recognition, showing incredible superiority to  conventional face recognition methods, such as EigenFace \cite{gupta2010new,rizon2006face,sahoolizadeh2008face} and subspace-based methods \cite{aishwarya2010face}. Bless by its over-parameterization nature, deep learning-based approaches enjoy unique advantages including (1) more powerful feature extraction for face representation, (2) an integrated representation and discriminative learning process in an end-to-end manner, and (3) higher capacity of memorization and generalization with large datasets~\cite{zhang2021rethink}.  

While deep face recognition approaches have been reported to outperform human's perception \cite{sun2014deep,sun2015deepid3}, they have been concerned with issues of reproducibility, \ie, it might be difficult to achieve the same performance when algorithms and models were re-implemented from the details released in the papers. Although some researchers release codes, it is still inconvenient to reproduce the experiments for fair comparisons as they often have been ``cooked with recipes'', \eg, tricks in architectures, losses, data processing, training and evaluation.
%
%
Hence, developing a comprehensive face recognition library, with all alternating backbones, loss functions and bag of tricks incorporated, is vital for both researchers and engineers. To this end, we develop a relatively comprehensive deep face recognition library named \emph{face.evoLVe} to meet the goals above.
In this paper, we present the features of the developed face.evoLVe library in details. 
In summary, our main contributions could be categorized in three folders as follows:
\begin{enumerate}
\item We develop a comprehensive library, namely face.evoLVe, for face-related analytics and applications, including face alignment (\eg, detection, landmark localization, affine transformation, \etc), data processing (\eg, augmentation, data balancing, normalization, \etc), where various backbones (\eg, ResNet \cite{he2016deep}, IR, IR-SE \cite{hu2018squeeze}, ResNeXt \cite{xie2017aggregated}, SE-ResNeXt, DenseNet \cite{huang2017densely}, LightCNN \cite{wu2018light}, MobileNet \cite{howard2017mobilenets}, ShuffleNet \cite{zhang2018shufflenet}, DPN \cite{chen2017dual}, \etc) with alternating losses (\eg, Softmax, Focal \cite{lin2017focal}, Center \cite{wen2016discriminative}, SphereFace \cite{liu2017sphereface}, CosFace \cite{wang2018cosface}, AmSoftmax \cite{wang2018additive}, ArcFace \cite{deng2019arcface}, Triplet \cite{schroff2015facenet}, \etc) and bags of tricks (\eg, training refinements, model tweaks, knowledge distillation \cite{hinton2015distilling}, \etc) for improving performance have been provided in standard implementations.

\item Face.evoLVe supports multiple popular deep learning platforms including both PaddlePaddle~\cite{ma2019paddlepaddle} and PyTroch \cite{paszke2019pytorch}. On top of the native platform, \ie, PyTorch, face.evoLVe provides necessary facilities to support parallel training with multi-GPUs, where users could enjoy the computation power of massive GPUs using few lines of codes/configurations. Note that the parallel training scheme in face.evoLVe not only supports the training of backbones~\cite{paszke2019pytorch}, but also accelerates the training of fully-connected (softmax) layers to fully scale-up the parallel training based on multi-GPUs and large datasets over distributed storage.


\item Face.evoLVe can help researchers/engineers develop high-performance deep face recognition models and algorithms quickly for practical use and deployment. Specifically, all data before and after alignment, source codes and trained models are provided, which reduces the efforts required for reproducing the existing methods, facilitates the development of new advanced approach, and provides training and evaluation environments for fair comparisons. We have used face.evoLVe to participate in a  number of face recognition competitions and secured the first place. In addition, the library is well designed and evolving vibrantly with a group of active contributors. New face recognition approaches can be easily plugged into the face.evoLVe framework.

\end{enumerate}


\section{Related Work}\label{sec2}
\subsection{A Brief Review of Face Recognition}
Normally, a face recognition system consists of face detection, facial landmark localization, face alignment, feature extraction and matching \cite{wang2018deep,jian2018deep}. 
Each part of the face recognition system can be an individual research area  and recently, a wide range of approaches have been proposed not only for the feature extraction module to obtain better representations of faces but also for other modules, such as  loss functions. Deepface \cite{taigman2014deepface} employs deep neural networks (DNNs), such as Alexnet \cite{krizhevsky2012imagenet} to extract face features, which is much more powerful than using Eigenface \cite{gupta2010new,rizon2006face,sahoolizadeh2008face}. A marginalized CNN is proposed by Zhao \etal \cite{zhao122017marginalized} to achieve more robust face representations. In terms of the loss function, Deepface \cite{taigman2014deepface} adopts softmax, which is widely used for classification \cite{krizhevsky2012imagenet,simonyan2014very,he2016deep}. In contrast, Sun \etal employ contrastive loss \cite{sun2015deepid3}. However, either softmax loss or contrastive loss is not sufficient to  learn discriminative features for face recognition. Also, Alexnet cannot obtain satisfactory representations of faces, hence triplet loss is applied in \cite{schroff2015facenet,liu2015targeting,parkhi2015deep} to learn more discriminative features. GoogleNet \cite{szegedy2015going} and VGGNet \cite{simonyan2014very} are then adopted to learn better representations. The problem of using triplet loss is that the training process is not stable. To mitigate this problem, Wen \etal \cite{wen2016discriminative} propose a center loss function, which learns a center for each class and penalizes the distances between the deep features and their corresponding class centers. More recently, angular loss functions \cite{wang2018cosface,deng2019arcface} and ResNet \cite{he2016deep} dominate face recognition. Researchers have realized that to accurately classify the queries, a face recognition model should strictly separate faces in the feature space. 
A number of approaches based on angular distance are proposed to achieve the goal, such as Cosface \cite{wang2018cosface}, Arcface \cite{deng2019arcface}, Regularface \cite{zhao2019regularface}, Adaptiveface \cite{liu2019adaptiveface} and Adacos \cite{zhang2019adacos}. 

Another challenging direction is age-invariant face recognition \cite{zheng2017cross,park2010age,zhao2020towards}, \ie, to learn representations of faces that is robust to appearance changes caused by facial aging. However, it is difficult to obtain sufficient well-annotated facial images with different age ranges, Zhao \etal \cite{zhao2019look} 
propose an age-invariant model to learn disentangled representations while synthesizing photorealistic cross-age faces for age-invariant face recognition.
Pose variants are also challenging for robust face recognition in the wild. 3D vision techniques have been used to estimate the pose and aid face recognition \cite{zhao20183d} under such settings.

In this paper, the developed face recognition library covers most state-of-the-art loss functions and backbones to achieve discriminative yet generative face representations for high-performance face recognition, which is also flexible for users to design their own loss function and backbone.

\subsection{Face Recognition Toolboxes}

\begin{table}[t]
    \centering
    \resizebox{\linewidth}{8mm}{
    \begin{tabular}{cccc}
    \toprule
         Library &\# backbones &\# heads &Platforms  \\ \hline
         \rowcolor{black!20} InsightFace \cite{insightface} &- &4 &Pytorch \cite{paszke2019pytorch}, MxNet \cite{chen2015mxnet} \\
         FaceX-zoo \cite{wang2021facex} &8 &8 &Pytorch \cite{paszke2019pytorch} \\
         \rowcolor{black!20} Face.evoLVe &13 &16 & Pytorch \cite{paszke2019pytorch}, PaddlePaddle \cite{ma2019paddlepaddle} \\
    \bottomrule
    \end{tabular}}
    \caption{Comparison between face.evoLVe and other libraries. InsightFace \cite{insightface} is the official implementation of Arcface \cite{deng2019arcface}.}
    \label{features}\vspace{-0.7cm}
\end{table}

Many widely used face recognition approaches have released their source codes, however, the platforms are different and the data processing method varies, which is inconvenient for the community of face recognition, resulting in unfair comparison. Therefore, using the same pipeline can be more convenient and flexible for face recognition. Guo \etal \cite{insightface} develop a toolbox named InsightFace for 2D and 3D face analysis, which supports two platforms: PyTorch \cite{paszke2019pytorch} and MxNet \cite{chen2015mxnet}. However, InsightFace only implements a few popular deep face recognition models, such as Arcface \cite{deng2019arcface} and Subcenter Arcface \cite{deng2020sub}. FaceX-Zoo \cite{wang2021facex} is a relatively comprehensive face recognition library, however, it only supports PyTorch platform \cite{paszke2019pytorch}, which is inconvenient for the users that use other platforms, such as Tensorflow \cite{abadi2016tensorflow} and PaddlePaddle \cite{ma2019paddlepaddle}.
By contrast, the proposed face.evoLVe library is highly flexible and scalable, which implements the complete face recognition pipeline, most face recognition models and supports both PyTorch and PaddlePaddle platforms. In addition, distributed training is well supported in face.evoLVe, which can be easily applied to large-scale datasets composed of millions of images. Tab. \ref{features} presents the comparison among different libraries, we can see that face.evoLVe is more comprehensive and supports more platforms. Another feature of face.evoLVe is that few-shot learning is supported.

\section{Face.evoLVe Library}\label{sec3}

\begin{figure}[t]
\centering
\includegraphics[width=\linewidth]{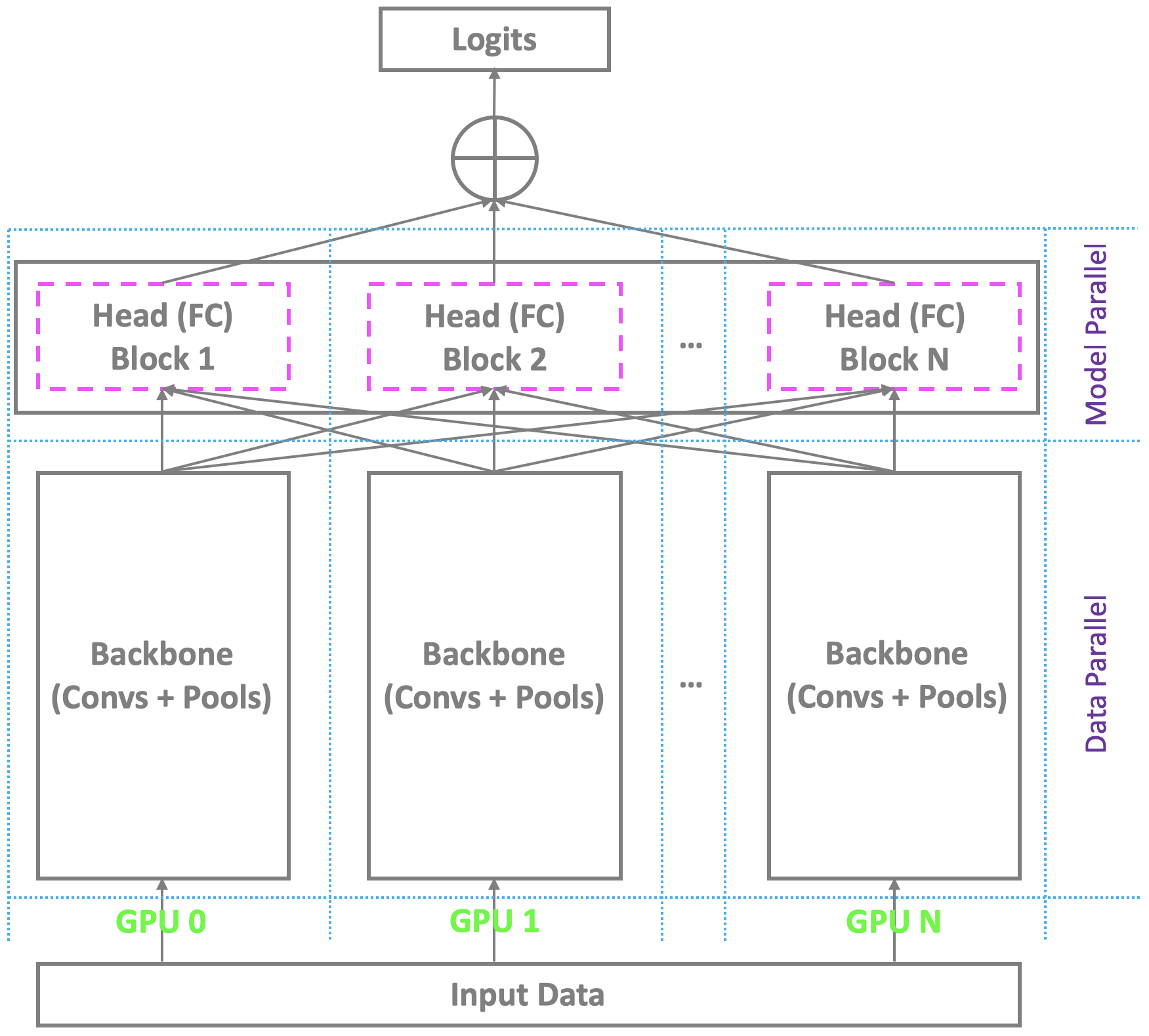}
\caption{An overview of the face.evoLVe pipeline. The input data can be images or videos. Face.evoLVe first detects faces in images or videos and then localizes the facial landmarks for alignment. The backbone network takes the processed data as input and outputs the representations of faces. Finally, the head blocks are used for compute the loss. Note that face.evoLVe supports distributed training to accelerate the training process.} \label{pipeline}
\end{figure}

In this section, we present the developed face.evoLVe library.
\subsection{Pipeline}
To be convenient and flexible, face.evoLVe library is well designed. We split the face recognition pipeline into four modules: face detection \& alignment, data processing, feature extraction and the loss head. Fig. \ref{pipeline} demonstrates the pipeline of the developed face.evoLVe library. face.evoLVe first detects faces in images and localize the facial landmarks for alignment, where MTCNN \cite{zhang2016joint} is adopted. The aligned faces are fed into the backbone networks to extract features, in face.evoLVe library, we implement a large number of backbones. Finally, face representations are used to compute the loss in the head block, where various loss function are implemented in the library. More details can be found in the following sections.

\subsection{Data Processing}
Given a dataset composed of $N$ images of $m$ classes, to balance the distribution of classes, face.evoLVe library first removes the low-shot classes that occurs less than \texttt{num\_min} times in the training dataset. It is easy for users to define their own \texttt{num\_min} for a specific dataset.

Apart from removing low-shot classes, face.evoLVe library also provides data augmentation approaches, \eg, flip horizontally, scale hue/satuation/brightness with coefficients uniformly drawn from $[0.6,1.4]$, add PCA noise with a coefficient sampled from a normal distribution $\mathcal{N}(0,0.1)$, \etc. Moreover, to alleviate the problem of long-tailed distribution, we also implement weighted random sampling in face.evoLVe library, which is user-friendly, \ie, users can also define a specific sampler instead of using the original weighted sampling implementation during training.

\subsection{Implemented Models}
One of the advantages of face.evoLVe library is that face.evoLVe contains many existing deep face recognition models. Generally, a model can be separated into two parts: (1) backbone -- to extract face features and (2) loss function -- to train a model and learn better representations of faces. Face.evoLVe library also designed in a highly-modular manner, \ie, backbones and loss heads are separately implemented.

\subsubsection{Backbone}
We implement many popular backbones in face.evoLVe library and users can easily change the configuration of backbones to specify the architecture of the backbone. In addition, the highly-modular design allows users to conveniently plug their own backbone networks into the library.

The implemented backbones in face.evoLVe library are as follows:
\begin{itemize}
    \item ResNet \cite{he2016deep} -- a deep convolutional neural network with residual connections, which has various versions, \eg, ResNet18, ResNet34, ResNet50, ResNet101 and ResNet152. ResNet has been widely applied to many vision tasks, such as classification \cite{he2016deep}, segmentation \cite{he2017mask} and object detection \cite{ren2015faster}, achieving state-of-the-art performance.
    \item IR -- an improved version of ResNet. Similar to ResNet, IR also has various versions with different numbers of layers.
    \item IR-SE -- applies squeeze-and-excitation (SE) blocks \cite{hu2018squeeze} to improve IR.
    \item ResNeXt \cite{xie2017aggregated} -- similar to ResNet, but splits channels into different paths, which achieves better performance on most vision tasks.
    \item SE-ResNeXt -- uses squeeze-and-excitation (SE) blocks \cite{hu2018squeeze} to improve ResNeXt.
    \item DenseNet \cite{huang2017densely} -- a deep convolutional neural network (CNN) with dense residual connections, \ie, a deep layer takes the outputs of all previous shallow layers as input, which takes more time.
    \item LightCNN \cite{wu2018light} -- adopts a Max-Feature-Map (MFM) for face recognition with noisy labels. It has 3 variants: LightCNN9, LightCNN29 and LightCNN29-V2, which is able to balance the performance and computational complexity.
    \item MobileNet \cite{howard2017mobilenets} -- a CNN with fewer parameters but achieves competitive performance compared with VGG16 \cite{simonyan2014very} that is 30 times larger.
    \item ShuffleNet \cite{zhang2018shufflenet} -- a very small CNN that uses pointwise group convolution, channel shuffle and depthwise separable convolution.
    \item DPN \cite{chen2017dual} --  a simple, highly efficient and modularized network. A shallow DPN surpasses the best ResNeXt101 \cite{xie2017aggregated} with smaller model size, less computational cost and lower memory consumption.
    \item AttentionNet \cite{yoo2015attentionnet} -- introduces attention mechanism \cite{bahdanau2014neural,vaswani2017attention} into ResNet \cite{he2016deep} to learn the attention-aware features.
    \item EfficientNet \cite{tan2019efficientnet} -- a series of models that consider width scaling, depth scaling, resolution scaling and compound scaling, resulting in lower volume size but comparable performance compared to the counterparts.
    \item GhostNet \cite{han2020ghostnet} -- a model that uses cheap operators to generate more features.
\end{itemize}

\subsubsection{Head and Loss Function}
A large number of heads and loss functions are implemented in face.evoLVe library, which covers most published works.

\begin{itemize}
    \item Softmax -- a simple and naive loss function for face recognition, which is popular in the earlier works.
    \item Focal \cite{lin2017focal} -- is used to alleviate the problem of class imbalance, which is first applied to object detection.
    \item Center \cite{wen2016discriminative} -- introduces intra-class distance into softmax, which forces the samples have the same class label are more close to each other.
    \item SphereFace \cite{liu2017sphereface} -- an angular softmax function. Compared to the original softmax function, Sphereface loss is more strict, hence the learned feature are more separable.
    \item CosFace \cite{wang2018cosface} -- cosine function and a large margin are applied to the original softmax function, therefore the samples are separated in the angular space.
    \item AmSoftmax \cite{wang2018additive} -- introduces a margin into angular softmax function to force the face recognition model to learn separable representations of faces. 
    \item ArcFace \cite{deng2019arcface} -- directly adds a margin to the angle, yielding a additive angular margin loss.
    \item Triplet \cite{schroff2015facenet} -- considers the relative difference of the distances among the triplet composed of an anchor image, a positive image and a negative image, which requires a triplet and how to select the negative images could be tricky.
    \item AdaCos \cite{zhang2019adacos} -- introduces dynamic scaling mechanism to adjust the margin and scale factor.
    \item AdaMSoftmax \cite{liu2019adaptiveface} -- adopts learnable margins and the average margin of all classes should be large.
    \item ArcNegFace \cite{liu2019towards} -- takes the distance between anchors into consideration to utilize hard negative mining and weaken the influence of the error labeling.
    \item CircleLoss \cite{sun2020circle} -- a general loss function, which is able to deduce other loss functions, such as triplet loss and AmSoftmax.
    \item CurricularFace \cite{huang2020curricularface} -- applies curriculum learning to face recognition and dynamically ranks the samples based on the hardness.
    \item MagFace \cite{meng2021magface} --  learns a universal feature embedding whose magnitude can measure the quality of the given face.
    \item NPCFace \cite{zeng2020npcface} -- adopts both hard negatives and positives to improve learning better representations.
    \item MVSoftmax \cite{wang2020mis} -- adaptively emphasizes the mis-classified feature vectors to guide the discriminative feature learning.
\end{itemize}

Note that focal loss \cite{lin2017focal} can be applied to all softmax-like functions, such as softmax, Arcface, and Cosface, which is able to mitigate the problem of class imbalance.

\subsection{Bag of Tricks}
Instead of implementing the existing works, such as backbones and loss functions, we also implement a bag of tricks for face recognition, which can be helpful for both researchers and engineers.

\textbf{Learning rate adjustment} -- at the beginning of the training process, all parameters are typically random values, therefore they are far away from the optimal. Using a large learning rate normally results in an unstable training process. One solution is using warmup \cite{gotmare2018closer}. In the warmup stage, we use a small learning rate in the beginning and then switch it to the initial learning rate when the training process is stable \cite{goyal2017accurate}. After warmup, we use learning rate decay -- the learning rate gradually decreases from the initial value. Cosine annealing strategy \cite{loshchilov2016sgdr} is also implemented in the library. An simplified version is reducing the learning rate from the initial value to 0 by following the cosine function. The cosine decay reduces the learning rate slowly in the beginning, and then becomes almost linear decreasing in the middle, and slows down again at the end. Compared to the step decay function, cosine decay is much more smooth and the learning rate is larger than using step decay in the middle training stage, resulting in faster convergence, which potentially improves the final performance.

\textbf{Label Smoothing} -- 
label smoothing was first proposed to train Inception-v2 \cite{szegedy2016rethinking}, which changes the labels from one-hot vectors to some distributions. We empirically compare the outputs of two ResNet50 models that are trained with and without label smoothing respectively and calculate the gap between the maximum output value and the average of the rest values. Under $\epsilon = 0.1$ and $K = 1000$, the theoretical gap is around 9.1 and using label smoothing, the output distribution center is close to the theoretical value and has fewer extreme values.

\textbf{Model tweak} -- a model tweak is a minor adjustment to the network architecture, such as changing the stride of a particular convolution layer. Model tweaks highly depend on the experience and knowledge of researchers. Generally, a tweak barely changes the computational complexity but might have a non-negligible effect on the model accuracy.  Many works have been done on ResNet tweak \cite{zhao2017pyramid,goyal2017accurate}. \Eg, changing the downsampling block of ResNet. Some tweaks \cite{hu2018squeeze,ma2018shufflenet,chen2017rethinking} replace the $7\times 7$ convolutional kernel in the input stem with stacking three conventional $3 \times 3$ convolutional kernels.

\begin{figure}[t]
\centering  
\includegraphics[width=0.48\linewidth]{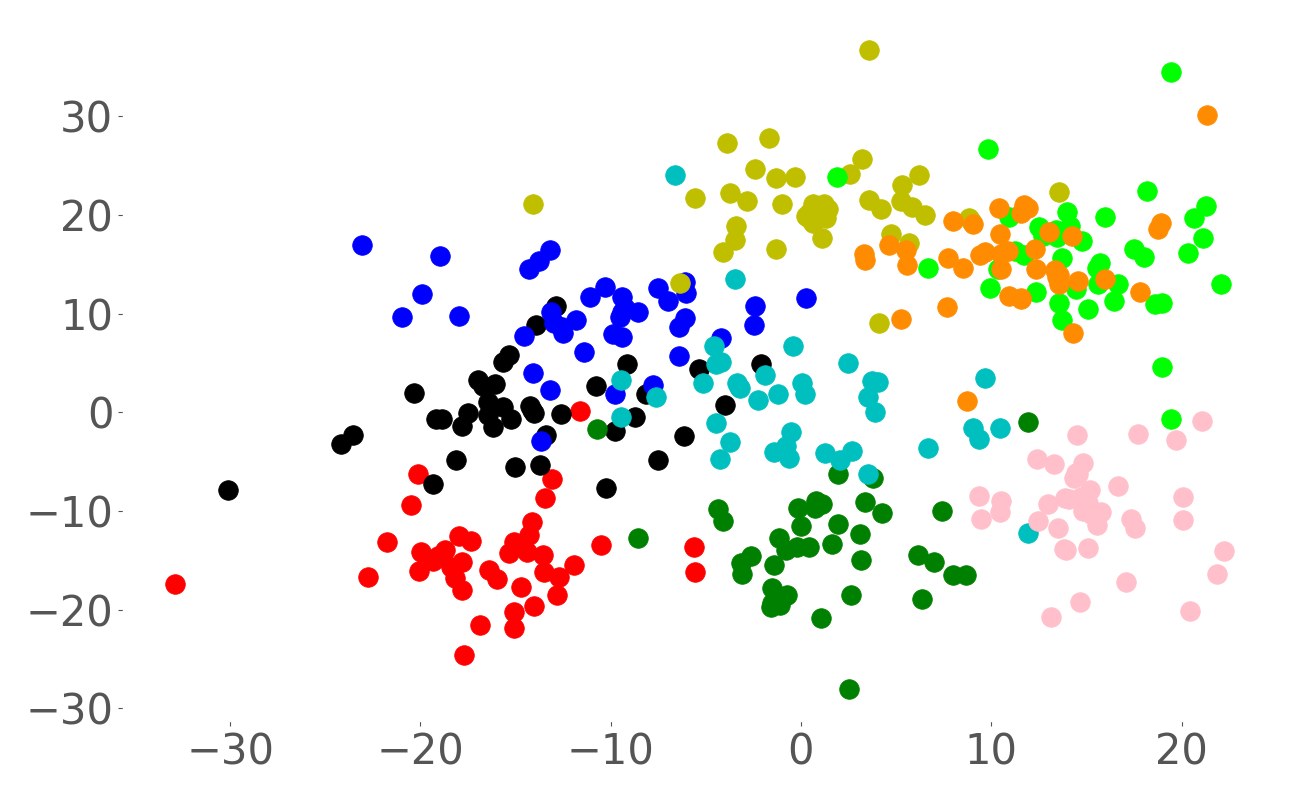}
\includegraphics[width=0.48\linewidth]{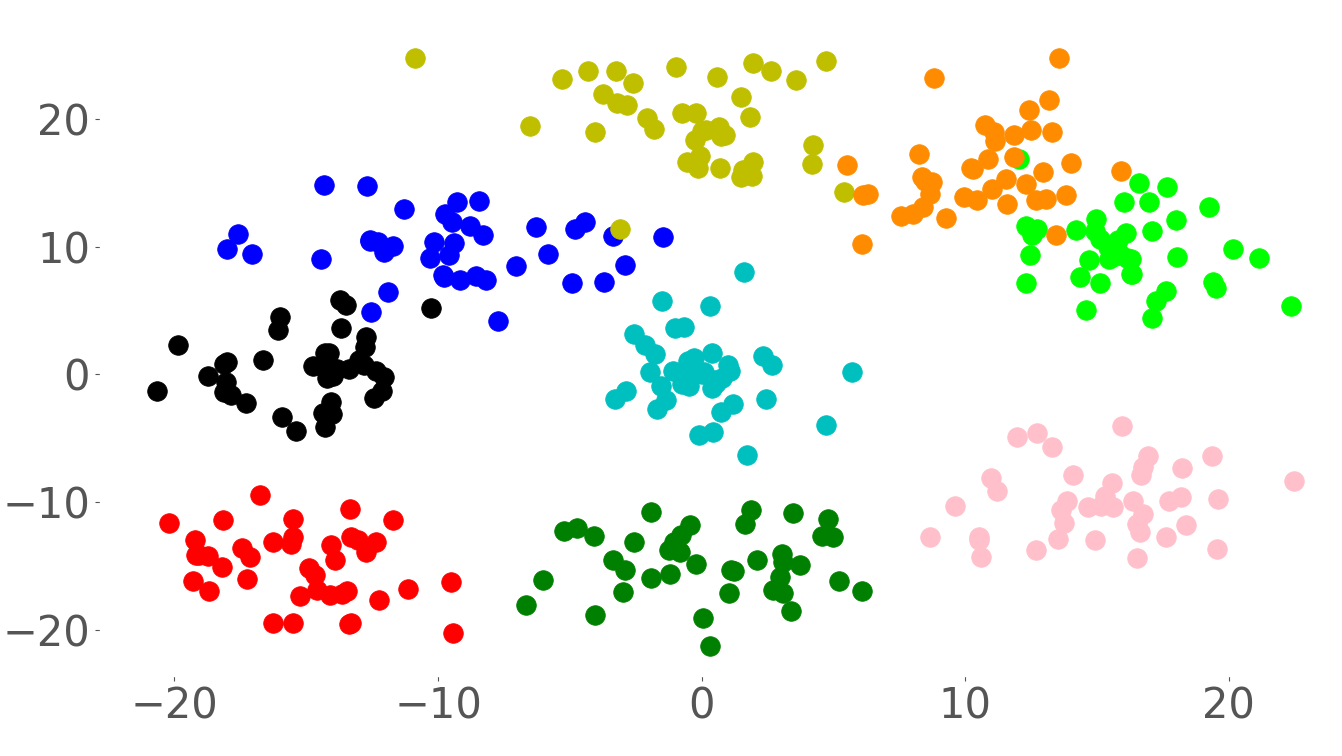}
\vspace{-1em}
\caption{T-SNE visualization of 10 classes randomly selected from MS-Celeb-1M. Left: without using knowledge distillation. Right: using knowledge distillation.}
\label{tsne-visualization}
\end{figure}

\textbf{Knowledge distillation} -- 
existing models suffer from noisy identity labels, such as outliers and label flips. It is beneficial to automatically reduce the label noise for improving recognition accuracy. 
Self-training is a standard approach in semi-supervised learning, which is explored to significantly boost the performance on image classification \cite{wu2018self,tai2021sinkhorn}. Based on our estimation, there are more than 30\% and 50\% noisy labels in MegaFace2 \cite{kemelmacher2016megaface,nech2017level} and MS-celeb-1M \cite{guo2016ms}. Since the datasets are relatively large and to reduce label noise and learning better representations of faces, we can apply self-training and knowledge distillation \cite{hinton2015distilling,kim2020self}, \ie, learning the knowledge of the previous trained model.In Fig. \ref{tsne-visualization}, we use T-SNE \cite{van2008visualizing} to visualize the distribution of the learned features with and without using knowledge distillation. Obviously, using knowledge distillation learns better representations since the features are more discriminative, \ie, it is much easier to classify the faces.

\subsection{Training and Evaluation}

\begin{table}[t]
    \centering
    \begin{tabular}{cccc}
    \toprule
         Dataset &\# identities &\# images  &\# videos \\ \hline
         \rowcolor{black!20} LFW \cite{huang2008labeled} &5,749	&13,233 &-\\
         CALFW \cite{zheng2017cross} &4,025 &12,174 &-\\ 
         \rowcolor{black!20} CASIA-WebFace \cite{yi2014learning} &10,575 &494,414 &-\\
         MS-Celeb-1M \cite{guo2016ms} &100,000 &5.08M &-\\ 
         \rowcolor{black!20} Vggface2 \cite{cao2018vggface2} &8,631 &3.09M &-\\ 
         AgeDB \cite{moschoglou2017agedb} &570 &16,488 &-\\ 
         \rowcolor{black!20} IJB-A \cite{klare2015pushing} &500 &5,396 &2,085\\ 
         IJB-B \cite{whitelam2017iarpa} &1,845 &21,798 &7,011\\ 
         \rowcolor{black!20} CFP \cite{sengupta2016frontal} &500 &7,000 &-\\ 
         Umdfaces \cite{bansal2017umdfaces} &8,277 &367,888 &-\\ 
         \rowcolor{black!20} CelebA \cite{liu2018large} &10,177 &202,599 &-\\ 
         CACD-VS \cite{chen2014cross} &2,000 &163,446 &-\\ 
         \rowcolor{black!20} YTF \cite{wolf2011face} &1,595 &- &621,127 \\ 
         DeepGlint \cite{deepglint} &180,855	&6.75M &-\\ 
         \rowcolor{black!20} UTKFace \cite{zhifei2017cvpr} &- &23,708 &- \\ 
         BUAA-VisNir \cite{huang2012buaa} &150 &5,952 &- \\ 
         \rowcolor{black!20} CASIA NIR-VIS \cite{li2013casia} 2.0 &725	&17,580 &-\\ 
         Oulu-CASIA \cite{zhao2011facial} &80 &65,000 &-\\ 
         \rowcolor{black!20} NUAA-ImposterDB \cite{tan2010face} &15 &12,614 &-\\ 
         CASIA-SURF \cite{zhang2020casia} &1,000 &- &21,000\\ 
         \rowcolor{black!20} CASIA-FASD \cite{zhang2012face} &50 &- &600\\ 
         CASIA-MFSD \cite{zhang2012face} &50 &- &600\\ 
         \rowcolor{black!20} Replay-Attack \cite{chingovska2012effectiveness} &50 &- &1200 \\ 
         WebFace260M \cite{webface260m} &2M &42M &-\\ 
    \bottomrule
    \end{tabular}
    \caption{Datasets supported by face.evoLVe library. All datasets are publicly available. And we also provide the pre-processed datasets.}
    \label{dataset} \vspace{-0.5cm}
\end{table}
Face.evoLVe supports most public datasets and users can directly download the datasets from the links provided in the library. Tab. \ref{dataset} presents the statistics of the datasets that supported by face.evoLVe library. For most datasets, we provide multiple versions, including the raw version and aligned version.

Note that large-scale datasets are more popular in recent years, hence distributed training is vital. Fortunately, face.evoLVe library well supports distributed training, thus users are able to train the model using relatively large datasets, such as MS-Celeb-1M \cite{guo2016ms} and WebFace260M \cite{webface260m}.

In terms of evaluation, we provide many trained models, so that researchers can easily evaluate the models for comparison and engineers can also quickly deploy a face recognition model.

\section{Performance}

\begin{table*}[t]
    \centering
    \resizebox{\textwidth}{23mm}{
    \begin{tabular}{cccccccccc}
    \toprule
         \multirow{2}{*}{Backbone} &\multirow{2}{*}{Head} &\multirow{2}{*}{Loss}  &\multicolumn{7}{c}{Testing Dataset} \\ 
         \cline{4-10}
         &&&LFW \cite{huang2008labeled} &CFP\_FF \cite{sengupta2016frontal} &CFP\_FP \cite{sengupta2016frontal} &AgeDB \cite{moschoglou2017agedb} &CALFW \cite{zheng2017cross} &CPLFW \cite{zheng2018cross} &Vggface2\_FP \cite{cao2018vggface2} \\
         \hline
         \rowcolor{black!20}
         IR50  &Arcface \cite{deng2019arcface} &Focal 
         &99.78 &99.69 &98.14 &97.53 &95.87 &92.45 &95.22 \\ 
         
         IR101 &Arcface \cite{deng2019arcface} & Focal  
         &99.81 &99.74 &98.25 &97.77 &95.93 &92.74 &95.44 \\ 
         \rowcolor{black!20}
         IR152 &Arcface \cite{deng2019arcface} & Focal 
         &99.82 &99.83 &98.37 &98.07 &96.03 &93.05 &95.50 \\ 

         IR50  &AdaCos \cite{zhang2019adacos} & Focal 
             &99.75 &99.53 &98.39 &97.25  &95.55  &92.25 &95.27 \\
         \rowcolor{black!20}
         IR101 &AdaCos \cite{zhang2019adacos} & Focal
             &99.78 &99.59 &98.41 &97.33  &95.68  &92.41 &95.35 \\
             
         IR152 &AdaCos \cite{zhang2019adacos} & Focal 
             &99.81 &99.65 &98.42 &98.47 &95.74 &92.57 &95.43 \\
         \rowcolor{black!20}
         IR50  &AM-Softmax \cite{wang2018additive} & Focal 
            &99.59 &99.59 &98.23 &97.37  &95.23 &92.37 &95.35 \\
            
         IR101 &AM-Softmax \cite{wang2018additive} & Focal 
            &99.65 &99.62 &98.31 &97.42  &95.38 &92.46 &95.47 \\ 
         \rowcolor{black!20}
         IR152 &AM-Softmax \cite{wang2018additive} & Focal 
            &99.72 &99.71 &98.45 &97.49  &95.50 &92.52 &95.55 \\
            
         LightCNN-9 &Arcface \cite{deng2019arcface} & Focal &98.75 &98.14 &92.57 &88.51 &89.23 &82.88 &92.14 \\
         \rowcolor{black!20}
         LightCNN-29 &Arcface \cite{deng2019arcface} & Focal &99.02 &98.57 &94.25 &90.85 &91.87 &85.78 &94.16 \\
         
         LightCNN-29v2 &Arcface \cite{deng2019arcface} & Focal &99.23 &98.47 &94.35 &91.49 &91.72 &85.78 &94.06 \\

    \bottomrule
    \end{tabular}}
    \caption{The performance of the implemented models in face.evoLVe library on different Head and Backbone. We train the models using Ms-Celeb-1M dataset \cite{guo2016ms}.}
    \label{performance}
\end{table*}

\begin{table*}[t]
    \centering
    \resizebox{\textwidth}{17mm}{
    \begin{tabular}{cccccccccc}
    \toprule
         \multirow{2}{*}{Backbone} &\multirow{2}{*}{Head} &\multirow{2}{*}{Loss} &\multicolumn{7}{c}{Testing Dataset} \\
         \cline{4-10}
         &&&  LFW \cite{huang2008labeled} &CFP\_FF \cite{sengupta2016frontal} &CFP\_FP \cite{sengupta2016frontal} &AgeDB \cite{moschoglou2017agedb} &CALFW \cite{zheng2017cross} &CPLFW \cite{zheng2018cross} &Vggface2\_FP \cite{cao2018vggface2} \\
         \hline
         \rowcolor{black!20}
         IR152 &AdaCos \cite{zhang2019adacos} & Focal &99.82 &99.84 &98.37 &98.07 &96.03 &93.05 &95.50 \\ 
         
         HRnet \cite{wang2020deep} &MV-Softmax \cite{wang2020mis} & Focal &99.82 &99.51 &98.41 &97.88 &95.43 &88.95 &94.70 \\ 
         \rowcolor{black!20}
         TF-NAS-A \cite{hu2020tfnas} &AM-Softmax \cite{wang2018additive} & Focal &99.82 &99.47 &98.33 &96.65 &94.32 &84.88 &91.38 
         \\ 
         
         GhostNet \cite{han2020ghostnet} &ArcFace \cite{deng2019arcface} & Focal &99.69 &99.52 &98.48 &97.29 &94.92 &85.25 &90.88 
          \\ 
         \rowcolor{black!20}
         AttentionNet \cite{yoo2015attentionnet} &AdaCos \cite{zhang2019adacos} & Focal &99.82 &99.47 &98.52 &96.89 &95.12 &87.23 &94.23 
          \\ 
         MobileFaceNet \cite{howard2017mobilenets} &AdaCos \cite{zhang2019adacos} & Focal &99.73 &99.84 &97.75 &95.87 &94.87 &89.29 &93.20
         \\ 
         
    \bottomrule
    \end{tabular}}
    \caption{The performance of the implemented models in face.evoLVe library on different Head and Backbone. We train the models using Web260M dataset \cite{webface260m}.}
    \label{performance2}
\end{table*}

\begin{figure*}[t]
    \centering
    \includegraphics[width=\linewidth]{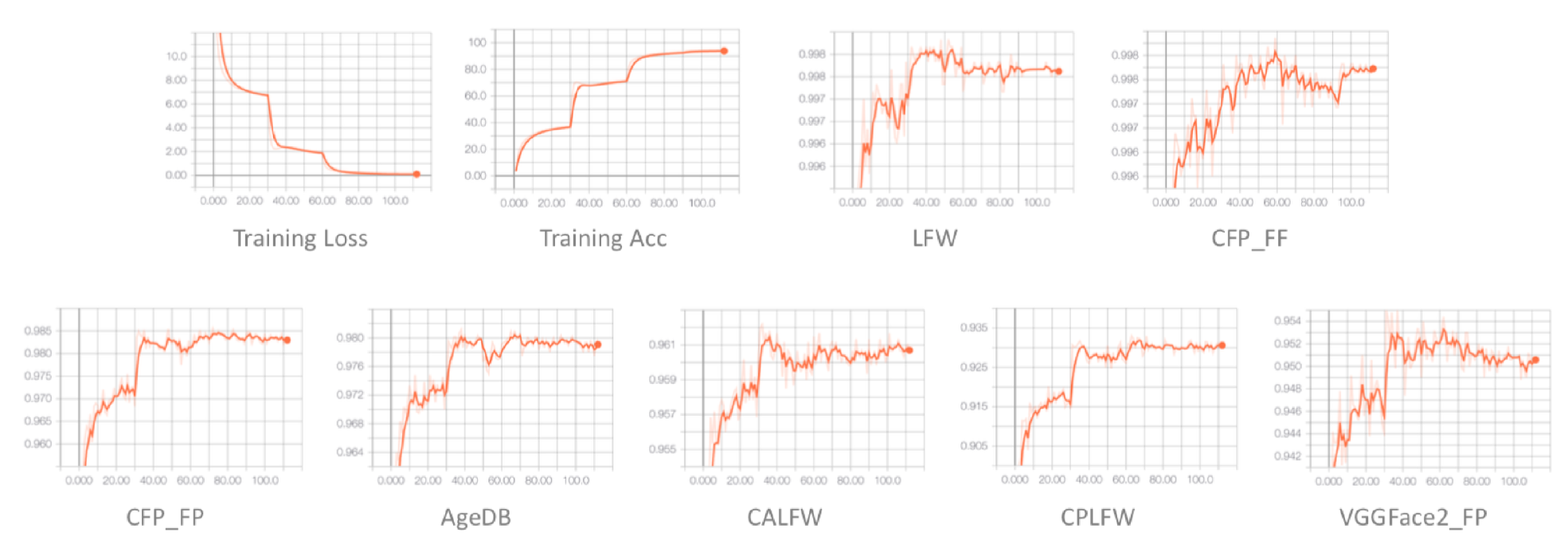}
    \vspace{-0.7cm}
    \caption{The visualization of the training process. We train the model on MS-Celeb-1M \cite{guo2016ms}, using IR152 as the backbone, Arcface \cite{deng2019arcface} as the head and focal loss. We validate the model on seven datasets and present the accuracy.}  \label{visualization}
\end{figure*}

We presents the performances of the implemented models in Tab. \ref{performance} and Tab. \ref{performance2}. We train the models using MS-Celeb-1M \cite{guo2016ms} dataset (Tab. \ref{performance}) and Web260M \cite{webface260m} (Tab. \ref{performance2}), then test them on different datasets. Compared with the state-of-the-art performance, the implemented models in face.evoLVe achieve competitive results, \eg, the original Arcface \cite{deng2019arcface} model with ResNet100 achieves 99.82 on LFW, 95.45 on CALFW and 92.08 on CPLFW, in contrast, using face.evoLVe with IR50 achieves 99.78 on LFW, 95.87 on CALFW and 92.45 on CPLFW (see tabel \ref{performance}), performing even better than the original Arcface. In addition, using deeper backbones like IR101 and IR152 are able to obtaining better performance. Compared to other implementations such as FaceX-zoo \cite{wang2021facex}, face.evoLVe is also competitive, \eg, using IR50+Arcface face.evoLVe and faceX-zoo achieve the same accuracy (99.78) on LFW, while face.evoLVe outperforms faceX-zoo on CALFW (95.87 vs. 95.47).

Interestingly, looking at Tab. \ref{performance2} where the models are trained on a relatively large dataset -- Web260M \cite{webface260m}, the performance could be further improved. \Eg, IR152+AdaCos trained on Web260M obtains 99.82, 96.03, 93.05 on LFW, CALFW and CPLFW, respectively, while the same model trained on MS-Celeb-1M achieves 99.81, 95.74 and 92.57.

Apart from presenting the accuracy, we also visualize the training process in Fig. \ref{visualization} to help users analyse the model behaviors during training, such as stability and overfitting. \Eg, we can easily find than overfitting occurs on LFW dataset during training, hence training more iterations does not improve the final performance but takes more time, in this case, early stop can be applied.

Moreover, we use face.evoLVe library to participate in many face recognition competitions, achieving the fist place and SoTA performance. \Eg, we achieve the fist place on ICCV 2017 MS-Celeb-1M Large-Scale Face Recognition Hard Set/Random Set/Low-Shot Learning Challenges by using the face.evoLVe library and No.1 on National Institute of Standards and Technology (NIST) IARPA Janus Benchmark A (IJB-A) Unconstrained Face Verification challenge and Identification challenge in 2017. In addition, face.evoLVe also achieves SoTA performance on other datasets, \eg, on MS-Celeb-1M hard set, using face.evoLVe obtains 79.10\% of coverage at precision=95\% and 87.50\% of coverage at precision=95\% on the random set.

\section{Using Customized Datasets and Models 
}

As we have mentioned, it is relatively easy and convenient to use the developed face.evoLVe library for both training and evaluation. Generally,  researchers also requires the modularity of a toolbox, thus they are able to easily plug their proposed models into the toolbox with changing only a few codes. Fortunately, face.evoLVe is a highly modular library. 

Apart from the datasets provided by the library, users can also their own datasets. Face.evoLVe provides data pre-processing SDK, including face detection and alignment, so that users can first use the pre-processing SDK to obtain the images of faces, which is relatively convenient.

In terms of the model, which is designed in a modular and extensible manner, \eg, the backbones and heads are independent to each other, thus, users can easily plug either customized backbones or heads without changing the architecture of the library. Also, during training users can easily change the hyper-parameters, such as learning rate, batch size and momentum.

\section{Conclusion}
In this paper, we have presented a comprehensive face recognition library -- \underline{face.evoLVe}, which is composed of necessary components covering the full pipeline of face recognition practices, including the alternating backbones and loss functions for face detection, alignment, feature extraction and matching. 
The goal of face.evoLVe is to lower the technical burden for researchers in the community to reproduce existing algorithms and models for comparisons and benchmark.
%
%
In addition, face.evoLVe is designed in a highly modular and extensible manner, where users could easily implement and plug their own models into the library for potential extension and contribution. Also, the developed library provides a bag of tricks to improve the performance and stabilize the training process. Note that we have used face.evoLVe to achieve SoTA performance and secured the first place for a series of open competitions.
Currently, face.evoLVe is still evolving with a group of active contributors. Commitments with novel models, tricks and datasets are welcome. 

\section{Acknowledgement}
This work was partially supported by the National Science Foundation of China 62006244.

\bibliographystyle{ACM-Reference-Format}
\bibliography{acmart}

\end{document}